# Language Free Character Recognition using Character Sketch and Center of Gravity Shifting

Masoud Nosrati *

Kermanshah Branch, Islamic Azad University, Kermanshah, Iran.

Fakhereh Rahimi

Kermanshah Branch, Islamic Azad University, Kermanshah, Iran.

Ronak Karimi

Kermanshah Branch, Islamic Azad University, Kermanshah, Iran.

**Abstract:** In this research, we present a heuristic method for character recognition. For this purpose, a sketch is constructed from the image that contains the character to be recognized. This sketch contains the most important pixels of image that are representatives of original image. These points are the most probable points in pixel-by-pixel matching of image that adapt to target image. Furthermore, a technique called "gravity shifting" is utilized for taking over the problem of elongation of characters. The consequence of combining sketch and gravity techniques leaded to a language free character recognition method. This method can be implemented independently for real-time uses or in combination of other classifiers as a feature extraction algorithm. Low complexity and acceptable performance are the most impressive features of this method that let it to be simply implemented in mobile and battery-limited computing devices. Results show that in the best case 86% of accuracy is obtained and in the worst case 28% of recognized characters are accurate.

## I. INTRODUCTION

Visual information is the most important type of information perceived, processed and interpreted by the human brain; recognition of patterns is automatically done and data are stored. But in the computer world, image processing covers a gamut of scopes such as computerized photography, medical/biological image processing, automatic character recognition, finger print and face recognition, pattern recognition, machine / robot vision [1,2]; and it is used for many applications like industrial applications and security applications [3-5] on various platforms and infrastructures like distributed systems and clouds [6,7]. Optical character recognition (OCR) is a process that gets an image as input and converts it typewritten or printed text with machine codes. There are many applications for OCR like data entry from printed paper data records, whether passport documents, invoices, bank statements, computerized receipts, business cards, mail, printouts of static-data, or any suitable documentation [8].

Significance of OCR is measured by accuracy of detection. Most of recent approaches need some datasets for training their classifiers. For example, in many studies like [9] and [10], neural networks are trained for recognizing the letters. A more advanced for of character recognition is handwriting recognition. It needs more precision in features extraction, because of diversity in the handwritings. Also, two general scopes are open for character recognition: offline and online character recognition [11]. In offline recognition, training is done before entering to the detection phase. So, it needs a previous knowledge about the characters. In online approach, characters are recognized at the same time of their appearance. Proposed methods in this approach have more real-time usages [12] and generally are utilized in handwriting recognition by predicting the next movements of pen on the pad to form a letter.

Despite the importance of accuracy, most of proposed methods have a high complexity. It is regarded as a negative property for these methods. Heuristic light complexity algorithms especially for the real-time usages have always been considered as critical issues in this area.

In this paper, we are going to propose a heuristic method for character recognition. This method is based on the creating a sketch of query images (images that contain a character to be recognized). The sketch partitions the black and white areas of image and specifies some points as the representatives of that segment. These points are chosen so that they are the most probable points in a pixel-to-pixel matching with pattern characters.

The proposed method can be utilized as an independent way for character recognition, or in combination of other classifiers for extracting the features of patters. One of the most important facets of this method is its language free recognition. In fact, it is based on the partitioning in the spatial domain of image and has nothing to do with the special features of characters in different languages. In fact, this method can even be used for similar shapes in an image, but with some modifications!

Results of experimental usage show the significance of our method independently. It shows that show that in the best case 86% of accuracy is obtained and in the worst case 28% of recognized characters are accurate.

In the follows of this paper, we will introduce our proposed method in the section 2. Also, a technique for coping with elongation of some letters with fonts like Times new roman is presented. Section 3 is dedicated to the experiments and results of applying the method to a set of 700 characters. Finally, conclusion is placed at the end of paper as the 4$^{th}$ section.

## II. PROPOSED METHOD

### 2.1 Creating the sketch of image

Our proposed method is based on the depiction of sketch of image. The term sketch is derived from rapidly executed freehand drawing that is not usually intended as a finished painting [13]. But in our work, sketch of an image is referred to a similar sized image which contains some of the basic features of initial one [14]. Sketch functions in spatial domain and it is related to the features of black and white areas. It is supposed that an image which contains the picture of a character has some special features, regarding the partition of space to black and white areas. For example, in letter 'A', space is partitioned to some black and white areas. 'A' always has 2 diagonal black areas in the sides. But, what makes this character dissimilar to 'A' with a different font is the width and the declination of them. Our proposed method looks for neutralizing such this differenced.

Let *QI* be the query image, which the features must be extracted. *Sketch(QI)* is a gray background identical sized image, that contains the features of black and white areas. Figure 1 shows a basic sketch for character 'A'. Sketch itself is consisted of 4 vertical and horizontal sections that are indicated in figure 2. In the follows, algorithm of extracting this sketch is presented.

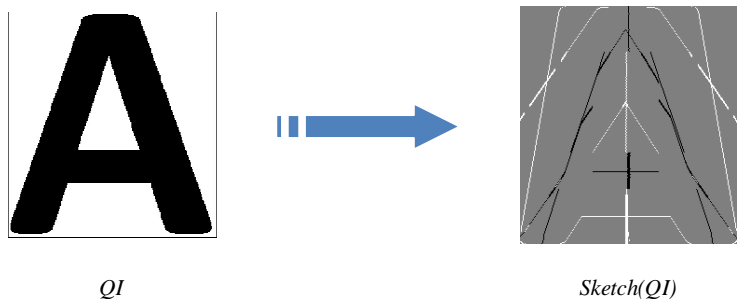

*QI*     *Sketch(QI)*

**Figure 1.** A basic example of sketch of character 'A'

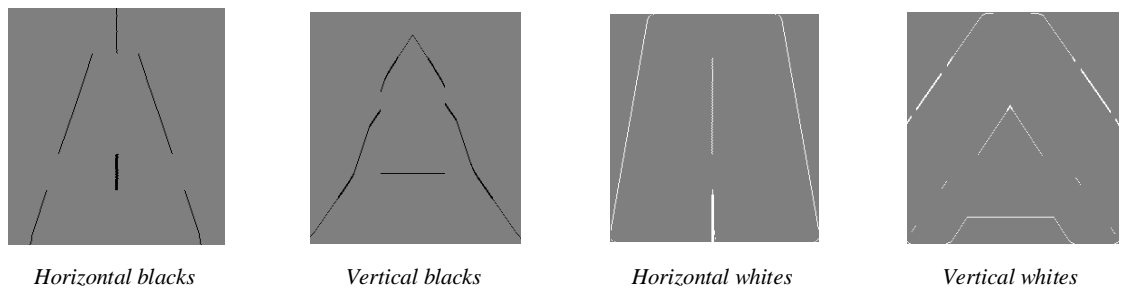

*Horizontal blacks*     *Vertical blacks*     *Horizontal whites*     *Vertical whites*

**Figure 2.** Construction of sketch

As it is seen in figure 1, *QI* contains some black and white areas. We aim to find the central points of each area. Main reason for doing so is that for adapting to different fonts, central points are more probable to be fitted. In other words, central points has the features of that area while adjusting the sketch to a new image. Figure 3 shows an example. As it is seen, most of the black points of sketch are dropped in the black areas and white points do so. Sketch can be utilized for training neural networks or any other classifier.

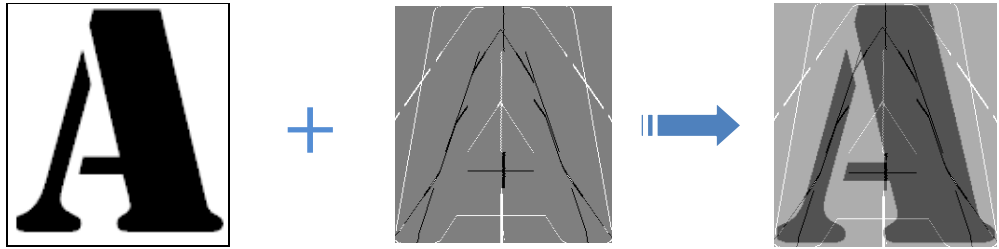

**Figure 3.** Adapting the black and white points of sketch to the sample image

As it is clearly seen, in the sketch of image all the areas are narrowed to 1 pixel width lines. In fact, in this model areas with wide width areas are turned to be 1 pixel width area as same as the areas with 1 pixel width. Emerging question is that "how can distinguish the wide areas from narrows in the sketch?"
To cope with the issue of diversified width, a pitch variable is considered, which we call it *bias*. Bias sets the width of black and white areas in the sketch image and it is as $0 < bias \leq 1$.
Effect of different bias parameter is shown in figure 4.

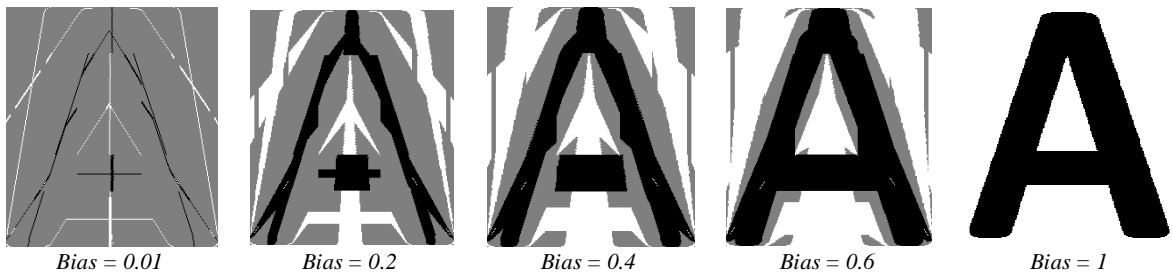

*Bias = 0.01*　　　*Bias = 0.2*　　　*Bias = 0.4*　　　*Bias = 0.6*　　　*Bias = 1*

**Figure 4.** Effect of different bias parameter on the sketch

For extraction of proposed sketch, following algorithm is presented. It is a sample for extraction of horizontal blacks as it was shown in figure 2. For horizontal whites and verticals the algorithm is the same, but instead of rows, columns must be explored and instead of black pixels, whites should be pointed.

### 2.2 Problem of elongations
There are many cases that the character has some extra elongated parts which cause problems in matching with the sketch. For example, suppose the sketch of letter 'E' with Arial font face, is used for recognition of 'E' with Time new roman font face. As it is observable, Times new roman adds extra elongation to fonts that causes the displacement of whole character and changing the white and black areas. Figure 5 shows this issue. Elongations of 'E' with Times new roman font face shifts the whole character to forward. It causes the mismatch of sketch of other font faces with it. So, a good idea is reducing the extra elongations from target characters. But how is it possible?

---

**Proposed algorithm for sketch extraction**

1. Input the *QI*.
2. Create an image with similar size of *QI* and gray background; then call it *Sketch(QI)*.
3. For all the rows in QI do the follows:

    3.1   For all the pixels of the row do:

   a) If current pixel is black, then store its location

   *// it is the initial of black area*

   b) If current pixel is the first white after the blacks, then:

   *//end of black area*

   - Calculate the *median point* of black area.
   - Calculate the *width of the area* in sketch by multiplying the width of black area in *bias*.
   - Point the calculated region with specified width on the *Sketch(IQ)* by black color.

4. Return the *Sketch(IQ)*.

---

The idea for coping with the elongations is using the gravity center location for modifying the position of whole character. The center of gravity of a distribution of pixels in space is the unique point where the weighted relative position of the distributed pixels sums to zero [15]. It is simply calculated in Matlab by 'Centroid' property of regionprops command. Center of gravity is indicated by a red point in figure 5. Center of QI is shifted to the right, because of features of 'E' with Arial font face. But, elongations of 'E' with Times new roman font face caused the center to be shifted backward. The distance between the centers of both images shows the amount of shift. Target image must be shifted by *d* where:

$$d = (d_x, d_y)$$
$$d_x = Center_{x_{Target}} - Center_{x_{QI}} \qquad \text{// shift in the x coordinate}$$
$$d_y = Center_{y_{Target}} - Center_{y_{QI}} \qquad \text{// shift in the y coordinate}$$

Matching of a *QI* to *Target* image is calculated by:

$$Accuracy = \frac{1}{2}\left(\frac{w}{N_w} + \frac{b}{N_b}\right) \times 100$$

Where:
$w$ is the number of white points of sketch that match to the white areas of target image.
$N_w$ is the whole number of white points of sketch.
$b$ is the number of black points of sketch that match to the black areas of target image.
$N_b$ is the whole number of black points of sketch.

It should be noted that this step is essential for matching process, but it is not necessary for the feature extraction process. On the other hand, the center can be calculated and stored as a metadata for further operations in the feature extraction phase.

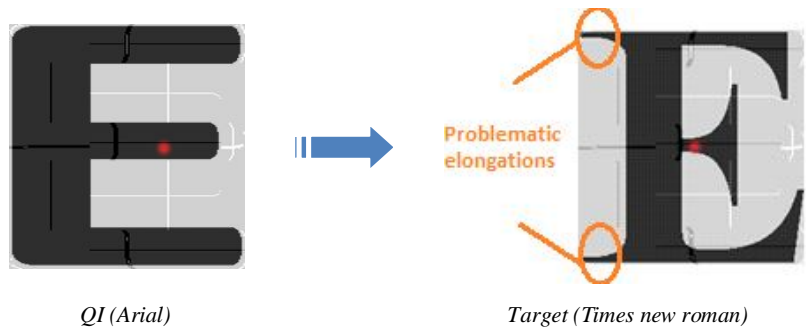

*QI (Arial)*          *Target (Times new roman)*

**Figure 5.** Mismatch of sketch of letter, caused by elongations of target image

### III. EXPERIMENTAL RESULTS

One of the most important goals for this study is to provide a low-level complexity program for optical character recognition. While utilizing some classifiers like neural networks is very useful for solving such these problems, they have a high complexity. Furthermore, the extracted features must be very precise. These types of classifiers are not suitable for real-time uses. So, low complexity methods are always been aid attention in this area. Even though our proposed method can be utilized for feature extraction as the input of a neural network, but we prefer to implement it as an independent method with single matches rather than thousands of matching to train a neural network or any other classifier. As another notion, the features of the letters in each language are different with other languages. It causes difficulties in designing multilingual methods for character recognition. The proposed method copes with this problem by partitioning the space and picking up the central points of each partition as the most probable point in matching. So, it makes this method a language free character recognition approach.

To evaluate the current method, we implemented a program in Matlab 2015a that created a list of Microsoft Windows standard fonts and tried to match the given query image to the characters of the fonts. Query images are called sample set. The sample set contains of 700 randomly created character with different font faces. A recognizer font is considered for detecting the letters of sample set. Each member of sample set is matched to all the characters of recognizer font. Table 1 shows the results of applying this method for detection of characters with different recognizer fonts.

The results show that in the best case 86% of accuracy is obtained and in the worst case 28% of recognized characters are accurate. As a further study, the combination of recognizer fonts can be investigated. Taking more into the experimental results is out of the focus of this paper, because obtained results show the significant of a low complexity method for character recognition.

**Table 1.** Results of proposed method with different recognizer fonts

| Sample sets | Recognizer font | Number of correct recognitions | Percent |
|---|---|---|---|
| Sample set containing randomly created 700 characters with different font faces | Arial | 587 | 83% |
| | Calibri | 473 | 67% |
| | Century | 489 | 69% |
| | Comic Sans MS | 398 | 56% |
| | Courier New | 311 | 44% |
| | Lucida console | 202 | 28% |
| | MS Gothic | 497 | 71% |
| | Tahoma | 567 | 81% |
| | Times new roman | 501 | 71% |
| | Verdana | 604 | 86% |

IV. CONCLUSION

In this research, we presented a new method for character recognition, based on construction of a sketch from the query image which contained a character to be recognized. This sketch contained the most important pixels of image that are representatives of original image. These points were the center of each black or white area and were the most probable points in pixel-by-pixel matching of image that adapt to target image, which caused mismatching. For pixel-to-pixel matching, there was a problem which was caused by elongation of characters by some font faces. For taking over this problem, center of gravity shifting technique was introduces. It neutralized the effect of elongations. The consequence of combining sketch and gravity techniques leaded to a language free character recognition method. This method can be implemented independently for real-time uses or in combination of other classifiers as a feature extraction algorithm. Low complexity and acceptable performance are the most impressive features of this method. Results show that in the best case 86% of accuracy is obtained and in the worst case 28% of recognized characters are accurate. For the future studies, integration of this method as a feature extraction algorithm with different classifiers for offline character recognition can be investigated.

REFERENCES

[1] M. Nosrati, R. Karimi, M. Hariri, K. Malekain, "Edge detection techniques in processing digital images: investigation of canny algorithm and Gabor method", World Applied Programming, vol. 3, pp. 116-121, March 2013.
[2] Masoud Nosrati, Ali Hanani Songhor and Koliaei, Ronak Karimi, "Steganography in Image Segments using Genetic Algorithm", Fifth International Conference on Advanced Computing & Communication Technologies, pp. 102-107
[3] Masoud Nosrati, Mehdi Hariri. An Algorithm for Minimizing of Boolean Functions Based on Graph DS, World Applied Programming, Vol (1), No (3), August 2011. 209-214.
[4] Sadeghi, M., Mohammadi, M., Nosrati, M., & Malekian, K. (2013). The Role of Entrepreneurial Environments in University Students Entrepreneurial Intention. World Applied Programming, 3(8), 361-366.
[5] Nosrati, M., Karimi, R, Mohamadi, M. y Maleikian, K. (2013). Internet Marketing or modern Advertising! How? Why? International Journal of Economy, Management and Social Sciences, 2 (3), 56-63
[6] Nosrati, Masoud, and Ronak Karimi. "Energy efficient and latency optimized media resource allocation." International Journal of Web Information Systems 12.1 (2016).
[7] Nosrati, Masoud, Abdolah Chalechale, and Ronak Karimi. "Latency Optimization for Resource Allocation in Cloud Computing System." Computational Science and Its Applications--ICCSA 2015. Springer International Publishing, 2015. 355-366.
[8] Schantz, Herbert F. (1982). The history of OCR, optical character recognition. [Manchester Center, Vt.]: Recognition Technologies Users Association. ISBN 9780943072012.
[9] John Resig (2009-01-23). "John Resig – OCR and Neural Nets in JavaScript". Ejohn.org. Retrieved 2013-06-16.
[10] Michael Sabourin1, Amar Mitiche. Optical character recognition by a neural network. Neural Networks, Volume 5, Issue 5, September–October 1992, Pages 843–852. doi:10.1016/S0893-6080(05)80144-3
[11] Huang, B.; Zhang, Y. and Kechadi, M.; Preprocessing Techniques for Online Handwriting Recognition. Intelligent Text Categorization and Clustering, Springer Berlin Heidelberg, 2009, Vol. 164, "Studies in Computational Intelligence" pp. 25–45.
[12] Nathan, K.S.; Beigi, H.S.M.; Subrahmonia, J.; Clary, G.J. (1995) Real-time on-line unconstrained handwriting recognition using statistical methods. Acoustics, Speech, and Signal Processing, 1995. ICASSP-95., 1995 International Conference on  (Volume:4 )
[13] Diana Davies (editor), Harrap's Illustrated Dictionary of Art and Artists, Harrap Books Limited, (1990) ISBN 0-245-54692-8
[14] Nosrati, M., Karimi, R., & Hasanvand, H. A. (2011). Extracting the Initial Sketch of Paintings Using Different Hough Transforms.
[15] Baron, Margaret E. (2004) [1969], The Origins of the Infinitesimal Calculus, Courier Dover Publications, ISBN 0-486-49544-2